\documentclass{article} 
\usepackage{iclr2027_conference,times}

\usepackage{amsmath,amsfonts,bm}

\def\eqref#1{equation~\ref{#1}}

\def\1{\bm{1}}

\DeclareMathAlphabet{\mathsfit}{\encodingdefault}{\sfdefault}{m}{sl}
\SetMathAlphabet{\mathsfit}{bold}{\encodingdefault}{\sfdefault}{bx}{n}

\usepackage{graphicx}

\usepackage{amsmath}
\usepackage{amssymb}
\usepackage{booktabs}
\usepackage{multirow}
\usepackage{graphicx}
\usepackage{subcaption}
\usepackage{wrapfig}
\usepackage[most]{tcolorbox}
\usepackage{amssymb}

\tcbset{aibox/.style={
    breakable,
    break at=\maxdimen,
    height fixed for=first and middle,
}}
\newcommand{\modelname}{HypoAgent}
\newtcolorbox{AIbox}[2][]{aibox,title={#2},#1}
\usepackage{hyperref}
\usepackage{url}

\title{ One Hypothesis Is Not Enough: \\ Abductive Reasoning with Agentic Hypothesis Refinement over Knowledge Graphs}

\author{%
\normalfont
\makebox[\dimexpr\textwidth-2\tabcolsep\relax][c]{%
\begin{tabular}{@{}c@{}}
\textbf{Yisen Gao$^{1}$},
\textbf{Yixi Cai$^{2}$},
\textbf{Tianshi Zheng$^{1}$},
\textbf{Jiaxin Bai$^{3}$},
\textbf{Yangqiu Song$^{1}$} \\[3pt]
$^{1}$The Hong Kong University of Science and Technology \\
$^{2}$Beihang University,
$^{3}$Hong Kong Baptist University \\[3pt]
{\small ygaodi@connect.ust.hk}
\end{tabular}}%
}

\iclrfinalcopy 
\begin{document}

\maketitle
\lhead{Preprint.}

\begin{abstract}
Abductive reasoning over knowledge graphs (KGs) seeks a first-order logic hypothesis whose answer set explains a given set of observed entities. Since many hypotheses can explain the same observations, controllable hypothesis generators condition generation on entities, relations, or logical patterns, but they treat generation as a single step. A generated hypothesis may be well-formed and satisfy the given conditions yet still fail to explain the observations, and the model has no mechanism to detect or correct this mismatch. Closing the abductive cycle requires revising a discrete, structured hypothesis, which large language models cannot do reliably, even though they iterate readily in natural language. We propose HypoAgent, an agentic hypothesis refinement framework that treats condition signals not only as expressions of user intent but also as operators that steer generation. A Hypothesis Proposal Agent calls a small trained generator to propose a hypothesis. Root-cause analysis then uses fragment-level coverage to locate branches worth inspecting and gathers neighborhood evidence from the training graph. A Hypothesis Refiner Agent combines this diagnosis with previously generated hypotheses to produce updated conditions and directly revised hypotheses. HypoAgent outperforms one-shot generation in single-turn and multi-turn settings on BioKG, PharmKG8k, and DBpedia50, and in the unconditional setting on DBpedia50. Our code is available at https://github.com/HKUST-KnowComp/HypoAgent.

\end{abstract}
\section{Introduction}

Abductive reasoning~\citep{abducreview} is one of the three classical modes of reasoning, alongside deductive~\citep{deductive} and inductive reasoning~\citep{inductive}. It reasons from evidence to hypotheses, seeking the most plausible explanation for observed phenomena. It is widely used in clinical diagnosis~\citep{medicaldianosis,medical2}, where physicians infer latent diseases or mechanisms from symptoms and patient profiles, and in scientific discovery~\citep{science1,science2,science3,science4}, where researchers propose and revise hypotheses to explain empirical observations.

Knowledge graphs (KGs) provide a natural substrate for abductive reasoning, as they represent domain knowledge as interconnected entities and relations. Over a KG, abductive reasoning~\citep{akgr} aims to find a first-order logic hypothesis whose answer set explains a given set of observed entities. For example, given several diseases, a system may hypothesize about their shared mechanisms, treatments, or susceptible populations. Such logical hypotheses are interpretable, since their answer sets and supporting paths can be inspected directly on the KG. However, the task is underdetermined: a single observation set may admit many plausible hypotheses, especially in large and densely connected KGs. Recent work~\citep{ctrlhgen} addresses this issue with controllable hypothesis generation, where users guide generation through explicit conditions, such as target entities, relations, or logical patterns, to obtain hypotheses that match their interests.

\begin{figure*}[t]
    \centering
    \begin{subfigure}[t]{0.49\textwidth}
        \centering
        \includegraphics[width=\linewidth]{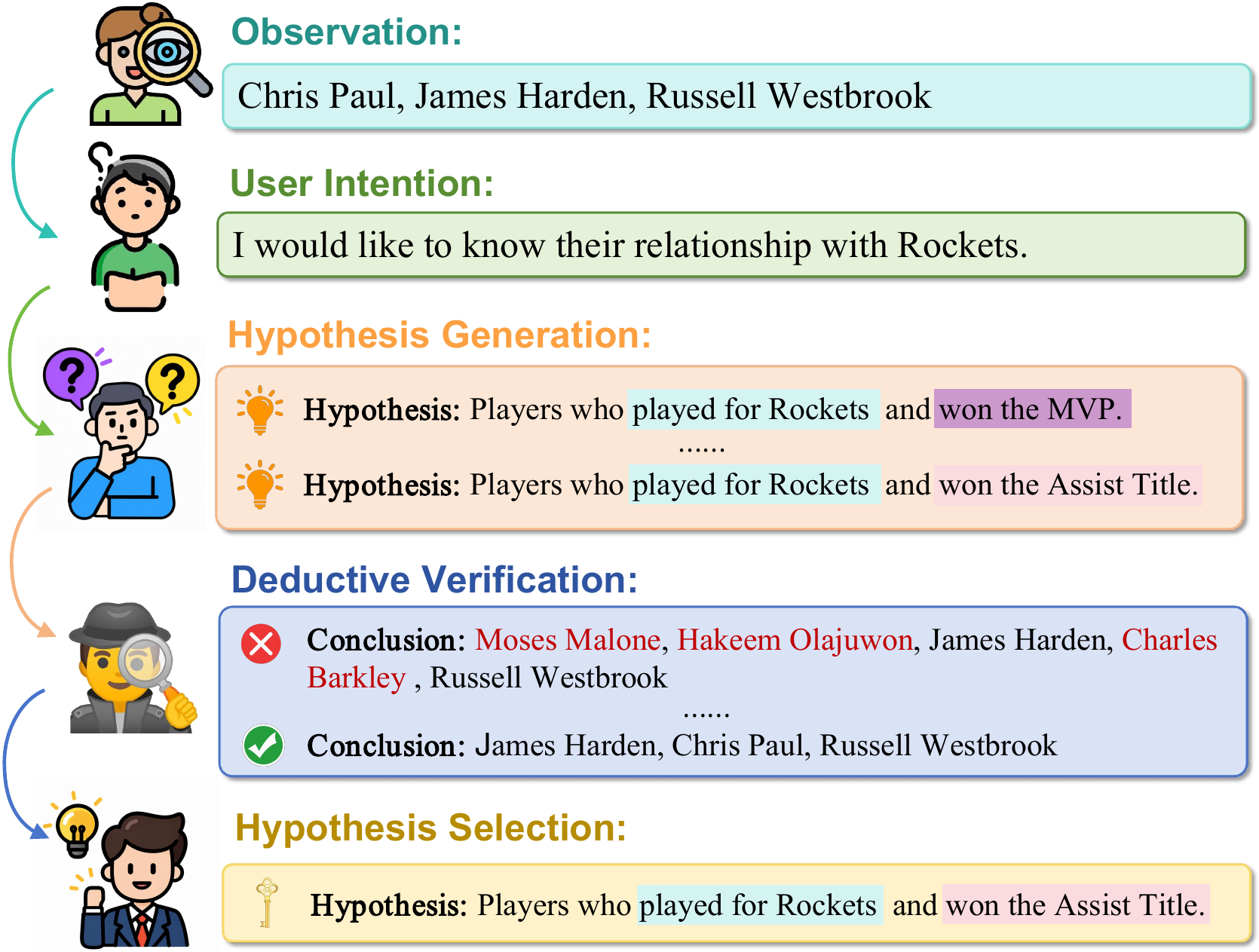}
        \caption{Two hypotheses that both satisfy the condition; only one explains the observations.}
        \label{fig:motivation-left}
    \end{subfigure}
    \hfill
    \begin{subfigure}[t]{0.49\textwidth}
        \centering
        \includegraphics[width=\linewidth]{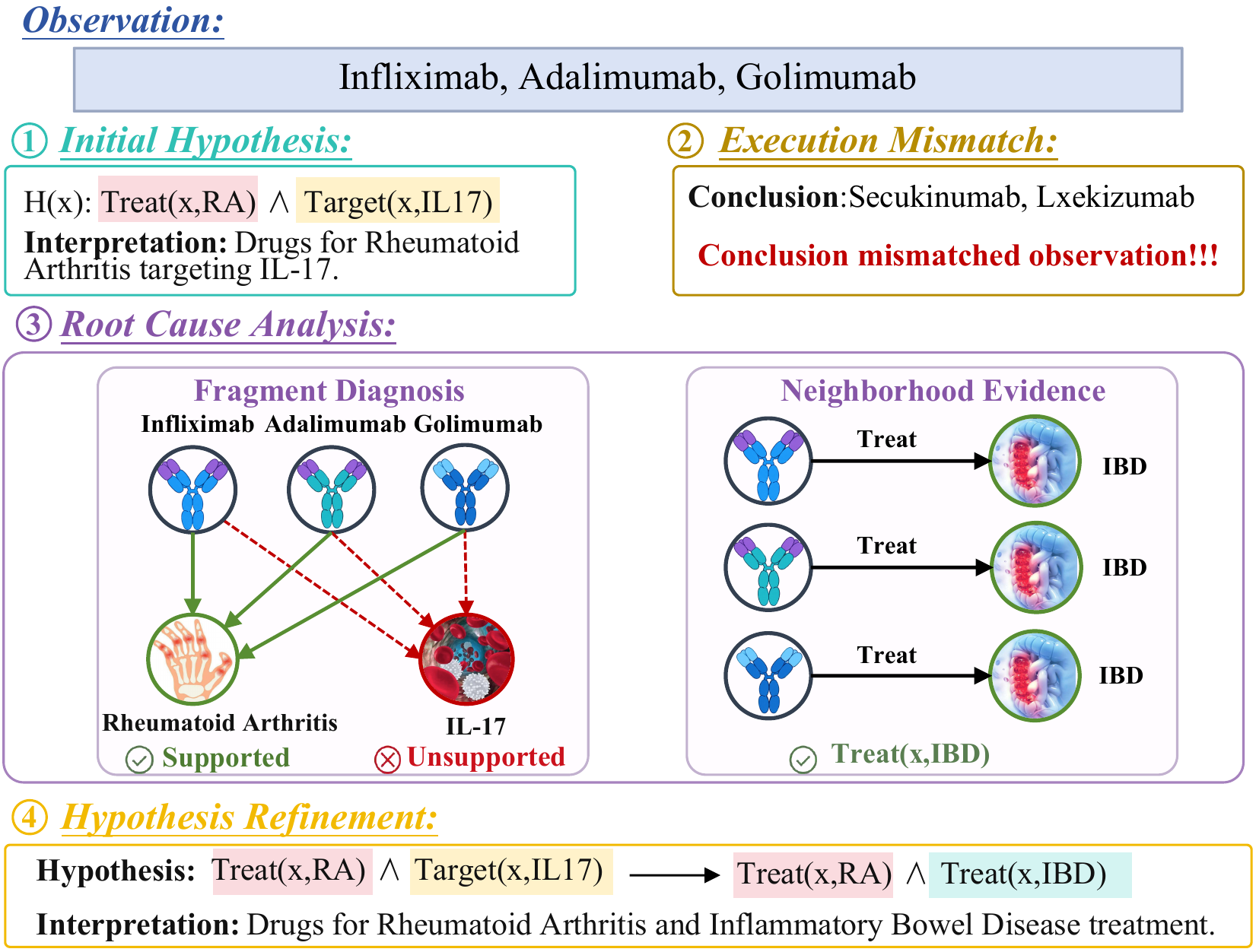}
        \caption{Fragment diagnosis and neighborhood evidence guide hypothesis repair.}
        \label{fig:motivation-right}
    \end{subfigure}
    \caption{Conditioning decides the direction of generation, not whether the hypothesis is correct.}
    \label{fig:motivation}
\end{figure*}

Abductive reasoning, however, is inherently iterative: hypotheses are proposed, validated, and refined. Controllable hypothesis generation~\citep{ctrlhgen} instead formalizes the task as one-shot conditional generation: given an observation set and a condition set, the model produces a hypothesis and terminates, executing only the first step of this cycle. A generated hypothesis may be well-formed and satisfy the given conditions, yet still explain the observations poorly, and there is no mechanism to detect or correct the mismatch. As shown in Fig.~\ref{fig:motivation-left}, given three NBA players as observations and a condition related to the Rockets, the model may hypothesize that they are players who played for the Rockets and won the MVP award. Although this hypothesis looks plausible and satisfies the condition, its answer set does not match the observed entities. Conditioning determines the direction of generation, but not whether the resulting hypothesis is correct.

Closing this cycle requires two capabilities. First, execution feedback should guide targeted revision rather than undirected regeneration: fragment-level coverage indicates which branches to inspect, and neighborhood evidence suggests alternative anchors. Fig.~\ref{fig:motivation-right} illustrates such a repair, where a target-based query is replaced by a treatment-based query that better matches the observations on the graph. Second, previous hypotheses should inform later attempts, since even an imperfect query may contain useful structures or anchors. Together, these capabilities call for an execution-guided search that reuses information from earlier candidates.

LLM agents can invoke tools and interpret execution feedback, but prior work shows that they struggle to generate strongly structured hypotheses~\citep{akgr,zheng2025logidynamicsunravelingdynamicslogical}. We therefore separate revision planning from structured hypothesis generation: the LLM expresses the intended change as conditions for a specialized generator, while direct rewriting remains available as a complementary route. This division of labor lets the LLM focus on deciding what to change and the generator on producing well-formed logical queries.

Building on this analysis, we propose \textbf{\modelname{}}, an agentic hypothesis refinement framework for abductive reasoning over KGs. Given a set of observed entities and a condition set, a Hypothesis Proposal Agent first invokes a small trained generator to produce an initial hypothesis. Root-cause analysis then examines this hypothesis at a fine-grained level from two complementary perspectives. Internally, the hypothesis is decomposed into executable fragments, whose coverage of the observations indicates which parts to inspect and which to reuse. Externally, the agent explores the KG neighborhoods of the observed entities to identify relations and entity anchors supported by the graph. The resulting diagnosis, together with previously generated hypotheses, is passed to a Hypothesis Refiner Agent, which decides how to revise the current hypothesis. Our key insight is that condition signals can not only encode user intent but also steer the direction of refinement. Accordingly, the Refiner both revises the logical form directly and produces modified conditions that encode the desired correction; these conditions are passed back to the Proposal Agent to obtain new structured proposals. Among the complete candidates, the system prioritizes adherence to the user conditions and then training-graph Jaccard similarity.

Our contributions are summarized as follows:
\begin{itemize}
    \item \textbf{Condition-mediated refinement.} We use execution feedback to turn a one-shot conditional generator into the proposal component of a refinement loop, with direct rewriting as a complementary route.
    \item \textbf{Root-cause analysis.} We combine fragment-level coverage statistics with KG neighborhood probing, guiding hypothesis revision with complementary internal and external evidence.
    \item \textbf{Empirical evidence.} \modelname{} outperforms one-shot generation in single-turn and multi-turn settings on BioKG, PharmKG8k, and DBpedia50, and in the unconditional setting on DBpedia50. On DBpedia50 single-turn tasks, it also requires fewer remote LLM calls and lower inference cost than ReAct-style agents.
\end{itemize}

\section{Related Work}

\paragraph{Abductive Reasoning.}
Abductive reasoning infers plausible explanations from observations and has been studied in a variety of general reasoning settings.
In natural language inference, $\alpha$-NLI~\citep{bhagavatula2020abductivecommonsensereasoning} framed abduction as a commonsense reasoning task in which systems select the most plausible explanation for an observed event sequence.
Subsequent studies improve abductive inference through decoding, prompting, self-consistency, and uncommon-sense reasoning~\citep{qin2021futureunsupervisedbackpropbaseddecoding,kadiķis2022embarrassinglysimpleperformanceprediction,chan2023selfconsistentnarrativepromptsabductive,zhao2024uncommonsensereasoningabductivereasoning}.
Beyond commonsense scenarios, abductive reasoning has also been explored in formal textual reasoning benchmarks such as ProofWriter~\citep{tafjord2021proofwritergeneratingimplicationsproofs}, open-world reasoning with LLMs~\citep{zhong2023chatablabductivelearningnatural,del2023truedetectivedeepabductive,thagard2024chatgptmakeexplanatoryinferences}, abstract logical reasoning~\citep{liu2024incompleteloopinstructioninference,zheng2025logidynamicsunravelingdynamicslogical}, and neuro-symbolic abductive learning~\citep{ABL,ABL2,ABL3}.

\paragraph{Abductive Reasoning on Knowledge Graphs.}
Abductive reasoning over knowledge graphs generates logical hypotheses that explain a set of observed entities or facts.
AbductiveKGR~\citep{akgr} first introduced this problem, formulating KG abduction as Transformer-based hypothesis generation in which a model produces candidate logical explanations for the observations.
CtrlHGen~\citep{ctrlhgen} then introduced controllable abductive hypothesis generation by conditioning the generator on user-specified control signals, so that generated hypotheses better satisfy the desired semantic or structural constraints.
DARK~\citep{dark} unified deductive and abductive reasoning over KGs by modeling their bidirectional relationship: it treats hypotheses and conclusions as two sides of a sequence and uses masked diffusion to reason in both directions.
Together, these efforts have advanced neural graph database reasoning~\citep{ngdb,ngdbench,ngdbzoo,kgpfn} by improving hypothesis generation and controllability and by better capturing user intent.

\section{Preliminary}

\paragraph{Notation.}
We define a knowledge graph as $\mathcal{G}=(\mathcal{E},\mathcal{R},\mathcal{T})$, where $\mathcal{E}$ is the set of entities, $\mathcal{R}$ is the set of relations, and $\mathcal{T}\subseteq \mathcal{E}\times\mathcal{R}\times\mathcal{E}$ is the set of observed triples.
A triple $(v_i,r,v_j)\in \mathcal{T}$ indicates that relation $r$ holds between entities $v_i$ and $v_j$.
Under the open-world assumption~\citep{openworld}, the KG is incomplete, and a missing edge does not imply that the corresponding real-world relation is false; we therefore evaluate queries by executing them on fixed graph snapshots.
In this work, a hypothesis \(H\) is represented as a first-order logic formula over the entities, variables, and relations in \(\mathcal{G}\), with existential quantifiers and logical connectives such as \(\wedge\), \(\vee\), and \(\neg\).
For example, a conjunctive hypothesis can be written as
\[
H(X)=\exists Z_1,\ldots,Z_k:\ a_1\wedge a_2\wedge\cdots\wedge a_m,
\]
where \(X\) is the target variable, \(Z_1,\ldots,Z_k\) are existentially quantified intermediate variables, and each \(a_i\) is a relational literal, i.e., either a relational predicate \(r(u,v)\) or its negation \(\neg r(u,v)\). Here, \(r\) is a relation in \(\mathcal{G}\), and \(u,v\) can be entities, the target variable \(X\), or existential variables from \(\{Z_1,\ldots,Z_k\}\).
Given a hypothesis $H$, we denote by $\mathcal{A}_{\mathcal{G}}(H)$ the answer set obtained by executing $H$ on $\mathcal{G}$.
That is, $\mathcal{A}_{\mathcal{G}}(H)\subseteq\mathcal{E}$ contains the entities that, when assigned to the target variable $X$, make $H(X)$ hold on $\mathcal{G}$.
Negation follows the standard graph-query semantics: $\mathcal{A}_{\mathcal{G}}(\neg H)=\mathcal{E}\setminus\mathcal{A}_{\mathcal{G}}(H)$.

\paragraph{Task definition.}
Given a knowledge graph $\mathcal{G}$ and a set of observed entities $\mathcal{O}=\{o_1,o_2,\ldots,o_n\}\subseteq\mathcal{E}$, abductive hypothesis generation over KGs aims to infer a plausible first-order logic hypothesis $H$ that explains the observations.
Let $\mathcal{H}$ denote the space of candidate logical queries. The quality of a hypothesis is measured by the similarity between its answer set and $\mathcal{O}$:
\begin{equation}
    H^\ast = \arg\max_{H\in\mathcal{H}}
\operatorname{sim}\left(\mathcal{A}_{\mathcal{G}}(H), \mathcal{O}\right),
\end{equation}

where $\operatorname{sim}(\cdot,\cdot)$ can be instantiated by set-based metrics such as the Jaccard similarity, Dice coefficient, and Overlap coefficient.
Since different hypotheses may share the same answer set, the goal is to find a hypothesis that explains $\mathcal{O}$, not to recover a unique ground-truth formula.
In controllable abductive hypothesis generation~\citep{ctrlhgen}, the model is additionally given a user-specified control condition \(\mathbf{c}\), and the goal is to generate a hypothesis \(H\) that both matches \(\mathcal{O}\) and satisfies \(\mathbf{c}\).

\section{Method}

\subsection{Overview}
\begin{figure*}[t]
    \centering
    \includegraphics[width=\textwidth]{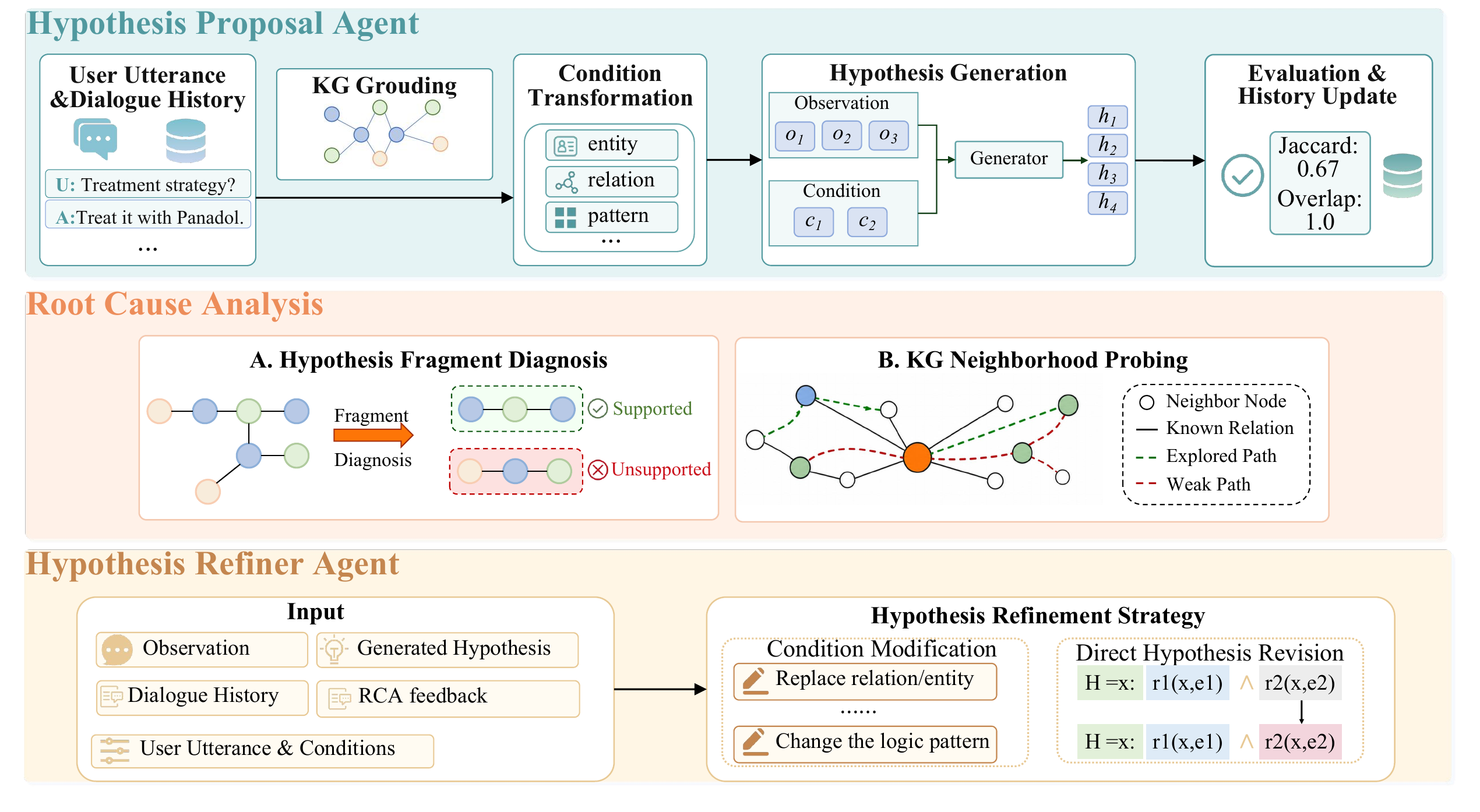}
    \caption{Overview of \modelname{}.}
    \label{fig:framework}
\end{figure*}

We propose \textbf{\modelname{}}, an agentic hypothesis refinement framework for abductive reasoning over knowledge graphs, as illustrated in Fig.~\ref{fig:framework}.
Given a knowledge graph, observations, and a user utterance, the system searches for a logical query that matches the observations while prioritizing the requested conditions.
The condition set is usually derived from the user's request; when none is given, the system induces proposal hints from an unconditional hypothesis (Appendix~\ref{app:inference}).
Each iteration consists of three sequential components.
First, a Hypothesis Proposal Agent (Section~\ref{sec:proposal}) calls a trained generator to propose a hypothesis from the observations and the extracted conditions, and validates it on the training graph. If the hypothesis is satisfactory, the loop terminates.
Otherwise, root-cause analysis (RCA; Section~\ref{sec:rca}) inspects the hypothesis at a fine-grained level and produces a diagnosis.
Finally, a Hypothesis Refiner Agent (Section~\ref{sec:refiner}) uses the diagnosis and previously generated hypotheses to produce revision candidates: a directly composed hypothesis and updated conditions for the Proposal Agent.

\subsection{Hypothesis Proposal Agent}
\label{sec:proposal}

Within a dialogue turn, $t$ indexes refinement rounds. At round $t$, the Proposal Agent generates a query under the current condition set $\mathbf{c}_t$, which may extend the user's conditions; the original user conditions stay fixed and serve as the selection criterion.
Following~\citet{ctrlhgen}, each condition set is a collection of key--value pairs $\mathbf{c}_t=\{(k_i,v_i)\}$, where each key $k_i$ is one of five types: \textit{relation}, \textit{entity}, \textit{relationnumber}, \textit{entitynumber}, and \textit{pattern}.
The first two specify semantic anchors, and the last three specify structural constraints on the target hypothesis.

To generate structured proposals efficiently, the Proposal Agent invokes a lightweight Transformer-based conditional generator $p_\theta$ instead of asking an LLM to write the logical form.
Given $\mathcal{O}$ and $\mathbf{c}_t$, the generator decodes a hypothesis $H_t$ autoregressively:
\begin{equation}
    p_\theta(H_t \mid \mathcal{O}, \mathbf{c}_t)
=
\prod_{\ell=1}^{L}
p_\theta(h^{\ell}_t \mid h^{<\ell}_t, \mathcal{O}, \mathbf{c}_t),
\end{equation}
where $h^\ell_t$ denotes the $\ell$-th token of $H_t$.
The logical hypothesis is then verbalized into natural language for interaction.

We train $p_\theta$ in two stages.
In the first stage, it is trained without conditions on sampled observation--hypothesis pairs to learn to generate well-formed hypotheses that explain $\mathcal{O}$.
In the second stage, it is trained on observation--condition--hypothesis triples to learn to satisfy the specified semantic and structural constraints while still explaining the observations.

At inference time, the Proposal Agent receives $\mathbf{c}_t$ (from the user, from a previous refinement round, or as an empty set) and calls $p_\theta$ to obtain $H_t$.
The hypothesis is executed on the training graph to obtain $\mathcal{A}_{\mathcal{G}_{\mathrm{train}}}(H_t)$, and its quality is measured by $\operatorname{sim}\bigl(\mathcal{A}_{\mathcal{G}_{\mathrm{train}}}(H_t),\mathcal{O}\bigr)$, instantiated as Jaccard similarity.
To avoid information leakage, all refinement signals are computed on $\mathcal{G}_{\mathrm{train}}$ rather than $\mathcal{G}_{\mathrm{test}}$.
If the score exceeds a threshold, $H_t$ is returned; otherwise, the system proceeds to root-cause analysis.

\subsection{Root-Cause Analysis}
\label{sec:rca}

Root-cause analysis uses execution feedback to investigate why $H_t$ fails to recover the observations.
It examines the failure from two complementary angles: the internal structure of $H_t$ and external evidence in the KG neighborhood of $\mathcal{O}$.

\textbf{Hypothesis Fragment Diagnosis.}
We first decompose $H_t$ into smaller executable fragments $\{H_t^{(1)},\ldots,H_t^{(m)}\}$.
Each fragment is a meaningful component of the formula, such as a single relational atom, an intermediate projection chain, or a partial conjunction.
The diagnostic tool executes these sub-queries on $\mathcal{G}_{\mathrm{train}}$ and reports their answer counts and overlap with the observations. For a fragment with answer set $A_t^{(j)}$, the overlap count is
\begin{equation}
    o_t^{(j)} = \left|A_t^{(j)} \cap \mathcal{O}\right|.
\end{equation}
The Refiner prioritizes low-overlap fragments for inspection and treats high-overlap fragments as candidate building blocks.
Since a fragment's role depends on its logical context (e.g., a branch with low standalone Jaccard can still be necessary for a correct intersection), these statistics guide revision, while complete hypotheses are always scored by training-graph Jaccard (Appendix~\ref{app:diagnosis}).

\textbf{KG Neighborhood Probing.}
In parallel, we probe the local neighborhood of the observed entities, where useful semantic anchors often appear.
The analysis searches for relations, entities, and short relational paths that frequently connect to the observations, and evaluates how well the corresponding candidate fragments recover $\mathcal{O}$.
High-scoring candidates can reveal relations or entities that are missing from the current hypothesis.

The diagnostic signal $\phi_t$ aggregates both views: fragment-level coverage statistics, branches to inspect, and candidate neighborhood anchors supported by $\mathcal{G}_{\mathrm{train}}$.

\subsection{Hypothesis Refiner Agent}
\label{sec:refiner}

The Hypothesis Refiner Agent decides how $H_t$ should be revised.
It receives the diagnostic signal $\phi_t$, the observation set $\mathcal{O}$, the user utterance $u_t$, and the history $\mathcal{D}_{\leq t}$ of previously generated hypotheses together with their Jaccard scores.
The key design is to use conditions as a structured repair interface: the diagnosis changes the generator's inputs, so the LLM need not express every repair as a logical query.
We therefore ask the Refiner both to revise the hypothesis directly and to propose updated conditions that steer subsequent generation toward a better match with the observations.
For example, on top of the user-intent conditions, it may add anchors suggested by the fragment-level coverage in $\phi_t$, or frequent relations and entities from the neighborhood of the observed entities.
If $u_t$ refers to a previous result (e.g., ``make it simpler''), it is interpreted with respect to $\mathcal{D}_{\leq t}$ rather than in isolation.
Formally, this process is written as
\begin{equation}
(\tilde{H}_{t+1},\, \mathbf{c}_{t+1})
=
\mathrm{Refiner}(\mathbf{c}_t,\, \phi_t,\, \mathcal{D}_{\leq t},\, u_t),
\end{equation}
where $\tilde{H}_{t+1}$ is the hypothesis revised by the Refiner and $\mathbf{c}_{t+1}$ is the updated condition set passed back to the Proposal Agent.
The Proposal Agent then uses $p_\theta$ to generate candidates under both the unchanged and the updated conditions, complementing the Refiner's direct revision.
Each candidate is executed on $\mathcal{G}_{\mathrm{train}}$ and ranked first by adherence to the user conditions and then by Jaccard similarity with $\mathcal{O}$.
The system keeps the best hypothesis across rounds and returns it once the similarity threshold is reached or the refinement budget is exhausted. Implementation details are given in Appendix~\ref{app:inference}.

\section{Experiment}

We evaluate \modelname{} along five research questions: single-turn controllable generation (RQ1), history-aware multi-turn refinement (RQ2), unconditional self-improvement (RQ3), component contributions (RQ4), and inference efficiency (RQ5).

\subsection{Experimental Setup}

\textbf{Datasets.}
We evaluate \modelname{} on three KG benchmarks: BioKG~\citep{biokg}, PharmKG8k~\citep{pharmkg}, and DBpedia50~\citep{dbpedia}.
Following prior work~\citep{akgr,ctrlhgen}, we partition KG edges in an 8:1:1 ratio and construct cumulative training, validation, and test execution graphs (Appendix~\ref{app:dataset}).
We sample hypothesis--observation pairs from 13 logical patterns and retain only test queries whose complete answer sets contain at most 32 entities, so no answer set is truncated.
The unconditional setting uses these sampled pairs directly.
For the single-turn and multi-turn dialogue settings, we further simulate user requests from the sampled cases: single-turn requests verbalize sampled control conditions, while multi-turn requests are built from cases that share the same observations but differ in hypotheses, reflecting progressive shifts in user intent.
Details are provided in Appendix~\ref{app:dataset}.

\textbf{Evaluation Metrics.}
We evaluate generated hypotheses in terms of semantic similarity and condition adherence.
For semantic similarity, we execute the selected hypothesis on $\mathcal{G}_{\mathrm{test}}$ and compare its answer set with the observation set $\mathcal{O}$ using the Jaccard similarity, Dice coefficient, and Overlap coefficient. All inference-time diagnosis, candidate validation, selection, and stopping use only $\mathcal{G}_{\mathrm{train}}$.
For condition adherence, Accuracy is the mean binary satisfaction of the dataset's gold conditions, which are independent of the conditions parsed by the model.

\textbf{Implementation Details.}
Following~\citet{ctrlhgen}, the hypothesis generator is a 6-layer Transformer.
For the LLM backend of the agents, we instantiate \modelname{} with DeepSeek-V4-Flash, Qwen3-235B~\citep{qwen3}, and GPT-5.4-mini, denoted as \modelname{}$_{\textsc{DS}}$, \modelname{}$_{\textsc{Qwen}}$, and \modelname{}$_{\textsc{GPT}}$, respectively. The maximum number of refinement rounds is set to three. All experiments are conducted on a single NVIDIA A6000 48GB GPU. Further details on training and agent construction are reported in Appendix~\ref{app:prompt}.

\begin{table*}[t]
    \centering
    \scriptsize
    \caption{Single-turn controllable hypothesis generation results. All values are means; across-case standard deviations are in Appendix~\ref{app:variance}. (\textbf{Bold}: best; \underline{Underline}: runner-up.)}
    \setlength{\tabcolsep}{2.3pt}
    \renewcommand{\arraystretch}{1.15}
    \begin{tabular}{lcccccccccccc}
    \toprule
    \multirow{2}{*}{\textbf{Model}} 
    & \multicolumn{4}{c}{\textbf{BioKG}} 
    & \multicolumn{4}{c}{\textbf{PharmKG8k}} 
    & \multicolumn{4}{c}{\textbf{DBpedia50}} \\
    \cmidrule(lr){2-5} \cmidrule(lr){6-9} \cmidrule(lr){10-13}
    & \textbf{Jaccard} & \textbf{Dice} & \textbf{Overlap} & \textbf{Accuracy}
    & \textbf{Jaccard} & \textbf{Dice} & \textbf{Overlap} & \textbf{Accuracy}
    & \textbf{Jaccard} & \textbf{Dice} & \textbf{Overlap} & \textbf{Accuracy} \\
    \midrule
    Symbolic Search
    & 0.420 & 0.480 & 0.623 & --
    & 0.495 & 0.583 & 0.771 & --
    & 0.460 & 0.485 & 0.539 & -- \\
    ReAct
    & 0.224 & 0.247 & 0.347 & 0.275
    & 0.216 & 0.248 & 0.343 & 0.315
    & 0.409 & 0.422 & 0.446 & 0.440 \\
    \midrule
    CtrlHGen
    & 0.718 & 0.761 & 0.858 & 0.909
    & 0.633 & 0.702 & 0.812 & 0.925
    & 0.781 & 0.829 & 0.896 & 0.730 \\
    CtrlHGen (best-of-$k$)
    & 0.758 & 0.803 & 0.895 & 0.904
    & 0.704 & 0.775 & 0.882 & 0.926
    & 0.810 & 0.852 & 0.912 & 0.710 \\
    ReAct + CtrlHGen
    & 0.718 & 0.768 & 0.893 & 0.802
    & 0.595 & 0.659 & 0.773 & 0.753
    & 0.839 & 0.873 & 0.934 & 0.676 \\
    \midrule
    \modelname{}$_{\textsc{DS}}$
    & \textbf{0.885} & \textbf{0.917} & \textbf{0.971} & \underline{0.941}
    & \textbf{0.826} & \textbf{0.858} & \textbf{0.947} & 0.935
    & \textbf{0.945} & \textbf{0.961} & \textbf{0.984} & \textbf{0.746} \\

    \modelname{}$_{\textsc{Qwen}}$
    & 0.859 & 0.890 & 0.951 & 0.940
    & \underline{0.817} & \underline{0.856} & 0.926 & \textbf{0.944}
    & 0.897 & 0.929 & 0.979 & 0.731 \\

    \modelname{}$_{\textsc{GPT}}$
    & \underline{0.860} & \underline{0.893} & \underline{0.953} & \textbf{0.945}
    & 0.812 & 0.855 & \underline{0.938} & \underline{0.943}
    & \underline{0.915} & \underline{0.946} & \underline{0.982} & \underline{0.736} \\
    \bottomrule
    \end{tabular}
    \label{tab:singleturn}
    \end{table*}
    
\begin{table*}[t]
    \centering
    \scriptsize
    \caption{Multi-turn hypothesis generation results. All values are means; across-case standard deviations are in Appendix~\ref{app:variance}. (\textbf{Bold}: best; \underline{Underline}: runner-up.)}
    \label{tab:multiturn}
    \setlength{\tabcolsep}{1.9pt}
    \renewcommand{\arraystretch}{1.15}
    \begin{tabular}{lcccccccccccc}
    \toprule
    \multirow{2}{*}{\textbf{Method}}
    & \multicolumn{4}{c}{\textbf{BioKG}}
    & \multicolumn{4}{c}{\textbf{PharmKG8k}}
    & \multicolumn{4}{c}{\textbf{DBpedia50}} \\
    \cmidrule(lr){2-5} \cmidrule(lr){6-9} \cmidrule(lr){10-13}
    & \textbf{Jaccard} & \textbf{Dice} & \textbf{Overlap} & \textbf{Accuracy}
    & \textbf{Jaccard} & \textbf{Dice} & \textbf{Overlap} & \textbf{Accuracy}
    & \textbf{Jaccard} & \textbf{Dice} & \textbf{Overlap} & \textbf{Accuracy} \\
    \midrule
    ReAct
    & 0.317 & 0.329 & 0.383 & 0.052
    & 0.279 & 0.294 & 0.326 & 0.062
    & 0.386 & 0.391 & 0.399 & 0.098 \\
    \midrule
    CtrlHGen (best-of-$k$)
    & 0.724 & 0.752 & 0.821 & 0.758
    & 0.815 & 0.856 & 0.931 & 0.766
    & 0.753 & 0.799 & 0.860 & 0.664 \\
    ReAct + CtrlHGen
    & 0.611 & 0.624 & 0.662 & 0.816
    & 0.659 & 0.678 & 0.731 & 0.773
    & 0.707 & 0.720 & 0.744 & \textbf{0.751} \\
    \modelname{} (w/o history)
    & \underline{0.892} & \underline{0.910} & \underline{0.956} & \textbf{0.847}
    & \underline{0.919} & \underline{0.940} & \underline{0.980} & \textbf{0.807}
    & \underline{0.895} & \underline{0.921} & \underline{0.957} & 0.704 \\
    \modelname{} (w/ history)
    & \textbf{0.910} & \textbf{0.926} & \textbf{0.959} & \underline{0.837}
    & \textbf{0.923} & \textbf{0.945} & \textbf{0.983} & \underline{0.805}
    & \textbf{0.910} & \textbf{0.934} & \textbf{0.970} & \underline{0.741} \\
    \bottomrule
    \end{tabular}
    \end{table*}

\subsection{RQ1: Single-turn Controllable Hypothesis Generation}

We compare \modelname{} with the following baselines in the single-turn setting.
\textit{Symbolic Search} performs structure-based search on the training KG.
Since it cannot interpret natural-language conditions, we use it only to find
the hypothesis with the highest semantic similarity and report its condition
accuracy as ``--''. \textit{ReAct} is a free-form ReAct-style LLM
agent~\citep{react} equipped with the same KG tools, namely \textit{Hypothesis Fragment Diagnosis}
and \textit{KG Neighborhood Probing}, but without the structured
refinement loop of \modelname{}. \textit{CtrlHGen}~\citep{ctrlhgen} is evaluated
with a single generation and with $k$ independent generations
(\textit{best-of-$k$}), where $k$ equals the maximum number of refinement
rounds of \modelname{}. The best-of-$k$ variant measures the gain obtainable
from repeated sampling alone, without diagnosis or refinement; note that $k$
matches the number of rounds rather than the total number of candidates.
Finally, \textit{ReAct + CtrlHGen} gives the same ReAct agent access to the
CtrlHGen generator as an additional tool.

Table~\ref{tab:singleturn} reports the results, with standard deviations in Appendix Table~\ref{tab:singleturn_std}.
First, the ReAct agent with KG tools performs poorly on structured hypothesis
generation and even falls behind symbolic search. This is consistent with prior
findings~\citep{ctrlhgen,dark} that even advanced general-purpose LLMs struggle
to understand and construct highly structured logical hypotheses.
Second, \textit{ReAct + CtrlHGen} is more competitive on DBpedia50, where
general world knowledge is more readily available, but remains weaker than
repeated CtrlHGen generation on the biomedical BioKG and PharmKG8k datasets.
A likely reason is that general-purpose LLMs are more familiar with the entities and relations in DBpedia50 than with those in biomedical graphs.
Third, \modelname{} achieves the best semantic similarity on all datasets and
metrics, substantially outperforming the baselines, while remaining competitive in
condition adherence. These gains hold across all three LLM backends, indicating that they do not depend on a particular LLM.
The comparison with CtrlHGen (best-of-$k$) is particularly informative: although both methods produce multiple candidates, \modelname{} achieves consistently higher similarity, showing that its gains come from diagnosis-guided refinement rather than repeated sampling alone.
\modelname{} also exhibits markedly lower across-case variability (Appendix Table~\ref{tab:singleturn_std}), suggesting that refinement mainly rescues cases in which the initial generation fails, rather than marginally improving cases that are already well explained.

\begin{figure*}[t]
    \centering
    \includegraphics[width=\textwidth]{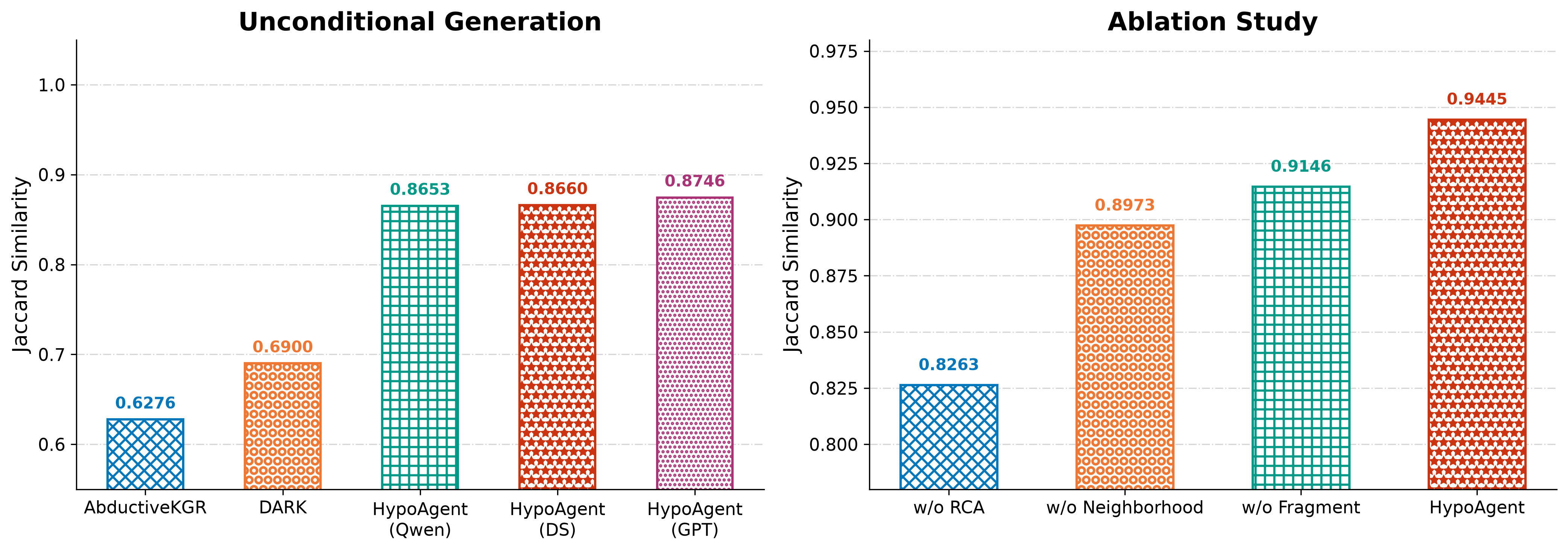}
    \caption{Jaccard similarity on DBpedia50 under the unconditional setting (left) and the single-turn ablation (right). Full ablation numbers are in Table~\ref{tab:singleturn_ablation}.}
    \label{fig:combined}
\end{figure*}

\begin{table}[t]
    \centering
    \scriptsize
    \caption{DBpedia50 single-turn inference usage, averaged per case, with DeepSeek-V4-Flash. Budgets are method-specific; counting details are in Appendix~\ref{app:budget}.}
    \label{tab:budget}
    \setlength{\tabcolsep}{3pt}
    \renewcommand{\arraystretch}{1.1}
    \begin{tabular}{lcccccc}
    \toprule
    \textbf{Method} & \textbf{Budget} & \textbf{Remote LLM}
        & \textbf{Generations} & \textbf{KG exec.} & \textbf{Cost (CNY)} & \textbf{Jaccard} \\
    \midrule
    ReAct & 10 steps & 9.52 & - & 4.93 & 0.1618 & 0.409 \\
    ReAct + CtrlHGen & 10 steps & 9.33 & 1.32 & 4.53 & 0.1440 & 0.839 \\
    \modelname{} & 3 rounds & \textbf{4.27} & 1.76 & \textbf{2.92} & \textbf{0.014} & \textbf{0.945} \\
    \bottomrule
    \end{tabular}
\end{table}

\subsection{RQ2: Multi-turn Goal-oriented Hypothesis Refinement}
Our multi-turn benchmark consists of scripted three-turn dialogues, where each turn targets a different reference hypothesis for the same observation set. Relative requests such as ``make the logic simpler'' can only be interpreted with conversational context. Since the dialogues are scripted, later requests do not adapt to each system's previous outputs.
We compare baselines that differ in how they use this history.
\textit{ReAct} is equipped with the same KG tools as in the single-turn setting.
\textit{CtrlHGen} receives conditions converted from the current user request by an LLM, and can be viewed as a variant of the Hypothesis Proposal Agent.
\textit{ReAct + CtrlHGen} uses CtrlHGen as a hypothesis generation tool; since it observes tool feedback, it also has access to previously generated hypotheses.
We further compare two variants of \modelname{} with DeepSeek-V4-Flash, which differ in whether previously generated hypotheses are provided as references for refinement.

Table~\ref{tab:multiturn} reports the results, with standard deviations in Appendix Table~\ref{tab:multiturn_std}.
In semantic similarity, ReAct agents, even with CtrlHGen as a generation tool, underperform CtrlHGen fed directly with the converted conditions, possibly because errors in earlier hypotheses carry over to later turns.
ReAct~+~CtrlHGen is more competitive on the commonsense DBpedia50 graph than on the biomedical BioKG and PharmKG8k graphs, and even achieves the highest condition accuracy on DBpedia50.
\modelname{} remains the strongest method overall: both variants outperform CtrlHGen and ReAct~+~CtrlHGen in semantic similarity on all three datasets. Its margin over CtrlHGen (best-of-$k$) widens relative to the single-turn setting on BioKG and DBpedia50, and history-aware refinement also yields higher condition accuracy than CtrlHGen on all three datasets, indicating that tracking earlier hypotheses helps the system follow evolving requests.
Providing previously generated hypotheses as references further improves semantic similarity on every dataset, e.g., raising BioKG Jaccard from 0.892 to 0.910, indicating that earlier hypotheses help guide subsequent refinement.
The effect on condition adherence is mixed: accuracy decreases slightly on BioKG (0.847 to 0.837) and PharmKG8k (0.807 to 0.805) but increases on DBpedia50 (0.704 to 0.741).

\subsection{RQ3: Unconditional Self-improving Hypothesis Generation}

We evaluate unconditional hypothesis refinement on DBpedia50, where no user-specified conditions are given.
Starting from an initial hypothesis, \modelname{} analyzes its execution results and the observed entities to induce conditions for subsequent generation.
We compare \modelname{} with AbductiveKGR~\citep{akgr} and DARK~\citep{dark},
representative autoregressive and diffusion-based abductive hypothesis
generation methods, respectively.

The left panel of Fig.~\ref{fig:combined} shows the Jaccard similarity on DBpedia50.
\modelname{} reaches 0.8653--0.8746 Jaccard across backends, compared with 0.6276 for AbductiveKGR and 0.6900 for DARK, showing that the framework remains effective without explicit user conditions.
The agent thus derives useful guidance from its own proposals and from graph evidence. Condition induction extends the refinement framework beyond user-directed generation, allowing the system to identify promising anchors and explore alternative hypotheses on its own.

\subsection{RQ4: Ablation Study}

The right panel of Fig.~\ref{fig:combined} shows the DBpedia50 single-turn ablation, and Table~\ref{tab:singleturn_ablation} reports Jaccard similarity on all three datasets.
Removing the entire RCA loop causes the largest drop, from 0.885 to 0.739 on BioKG, from 0.826 to 0.644 on PharmKG8k, and from 0.945 to 0.826 on DBpedia50,
confirming the benefit of execution-guided iterative refinement over one-shot conditional generation.
Removing neighborhood probing hurts more than removing fragment diagnosis on PharmKG8k and DBpedia50, while the two ablations perform nearly identically on BioKG.
The full model performs best on every dataset, reflecting the complementary roles of the two components: fragment diagnosis guides revision of the current hypothesis, while neighborhood probing supplies alternative relations and anchors.

To examine the two revision routes, we remove each candidate category from the recorded candidate pools and reselect the best remaining hypothesis (Appendix~\ref{app:candidate-ablation}); regeneration under unchanged conditions remains available in both settings. Removing condition-update candidates reduces Jaccard by 0.128--0.148, compared with 0.032--0.091 for direct rewriting. The larger drop shows that condition updates contribute beyond repeated generation under the same constraints, while the smaller but consistent drop for direct rewriting shows the value of keeping both routes.
At the sample level, removing condition-update candidates lowers test Jaccard for 44.2\%, 53.0\%, and 33.6\% of cases on BioKG, PharmKG8k, and DBpedia50, respectively, compared with 14.0\%, 15.6\%, and 24.0\% when direct rewriting is removed. Both routes thus provide useful alternatives for a substantial fraction of cases.

\subsection{RQ5: Inference Efficiency}

Table~\ref{tab:budget} compares \modelname{}$_{\textsc{DS}}$ with the two ReAct agents on DBpedia50 under their respective
inference budgets, using DeepSeek-V4-Flash as the LLM backend for all methods.
Compared with ReAct~+~CtrlHGen, \modelname{}$_{\textsc{DS}}$ achieves higher Jaccard similarity with
$2.2\times$ fewer remote LLM calls and about $10.3\times$ lower cost:
it raises Jaccard from 0.839 to 0.945 while reducing the average number of remote calls from 9.33 to 4.27 and the cost from 0.1440 to 0.0140 CNY per case. It also requires fewer KG tool executions, since diagnosis is invoked only when the current hypothesis falls below the threshold. Hence, giving a free-form agent access to the same generator does not by itself achieve the quality--cost trade-off of structured refinement; combining targeted condition updates with specialized generation yields stronger hypotheses at lower remote inference cost. Detailed accounting is provided in Appendix~\ref{app:budget}.

\subsection{Case Study}

Appendix~\ref{app:case} presents three examples of how \modelname{} turns diagnosis into revision: targeted anchor updates on PharmKG8k, history-guided structural changes on BioKG, and condition induction without user guidance on DBpedia50.
In the PharmKG8k example, adding the neighborhood-derived anchor DB11914 to the conditions yields a two-branch hypothesis whose Jaccard rises from 0.231 to 1.000, whereas directly combining this anchor with an anchor retained from the initial hypothesis scores only 0.083. The example highlights the role of the generator beyond retrieval: a promising anchor must be composed with a suitable partner, and condition updates let the generator search for this composition instead of requiring the LLM to specify it.

\section{Conclusion}

We proposed \modelname{}, an agentic framework for iterative abductive hypothesis generation over knowledge graphs.
Its central idea is to use conditions as an interface for revision: an LLM translates execution feedback and neighborhood evidence into updated conditions, and a specialized generator turns this guidance into structured hypotheses.
Direct rewriting complements condition-based generation, and previous hypotheses inform subsequent refinement.
Experiments show improved semantic similarity in single-turn and multi-turn generation on BioKG, PharmKG8k, and DBpedia50, as well as in unconditional generation on DBpedia50. On DBpedia50, \modelname{} also achieves higher quality than ReAct-style agents with fewer remote LLM calls and lower cost. Together, these results demonstrate the value of coupling agentic revision planning with specialized hypothesis generation to move beyond one-shot abduction.



\clearpage
\bibliography{iclr2027_conference}
\bibliographystyle{iclr2027_conference}

\appendix
\section{Data Construction}
\label{app:dataset}

KG edges are partitioned into disjoint training, validation, and test subsets in an 8:1:1 ratio. Execution graphs are cumulative: $\mathcal{G}_{\mathrm{train}}$ contains training edges, $\mathcal{G}_{\mathrm{valid}}$ adds validation edges, and $\mathcal{G}_{\mathrm{test}}$ contains all three subsets.

Following the sampling protocol of CtrlHGen~\citep{ctrlhgen}, a query is admitted to the test set only if its full test-graph answer set contains at most 32 entities; answers are neither truncated nor subsampled. Test cases are constructed so that the target hypothesis has additional answers on the cumulative test graph that are absent from its answer sets on the training and validation graphs. The test graph is used only for data construction and final evaluation, never for candidate selection or refinement at inference time.

\paragraph{Evaluation target.}
For a sampled reference query $H_{\mathrm{ref}}$, the observations are its complete test-graph answer set, $\mathcal{O}=\mathcal{A}_{\mathcal{G}_{\mathrm{test}}}(H_{\mathrm{ref}})$, with $|\mathcal{O}|\leq32$. Final semantic evaluation compares $\mathcal{A}_{\mathcal{G}_{\mathrm{test}}}(\hat H)$ with $\mathcal{O}$, so the reference query attains a Jaccard of 1. Any alternative explanation with the same answer set is equally rewarded, and syntactic recovery of $H_{\mathrm{ref}}$ is not required. Since $\mathcal{O}$ contains answers that the reference query cannot retrieve on the earlier graph splits, the training-graph feedback used during inference differs from the final test-graph score.

\paragraph{Condition labels.}
Reported Accuracy is computed against the dataset's gold conditions, independently of the model's condition parsing. Conditions parsed during inference guide proposal generation and candidate ranking but are not used as evaluation labels, so parsing errors are penalized rather than hidden.

\paragraph{Hypothesis sampling.}
Given a knowledge graph \(G\) and a predefined logical pattern \(P\), the algorithm samples a random node \(v\) as the target answer and recursively instantiates a hypothesis whose answer set contains \(v\) and whose structure matches \(P\).
At each step of the recursion, the current logical operator determines how the hypothesis is expanded.
For projection, the algorithm samples an incoming edge \((u,r,v)\) of \(v\) and recursively constructs the preceding sub-hypothesis from \(u\).
For intersection, all sub-hypotheses are instantiated with the same target node \(v\), since \(v\) must satisfy every branch.
For union, one branch is instantiated to derive \(v\) and the remaining branches are instantiated from randomly sampled nodes, since satisfying any single branch suffices for \(v\) to be in the union.

We use only the original directed edges of the knowledge graph and do not add inverse edges.
\begin{wrapfigure}{r}{0.48\textwidth}
    \centering
    \vspace{-8pt}
    \includegraphics[width=\linewidth]{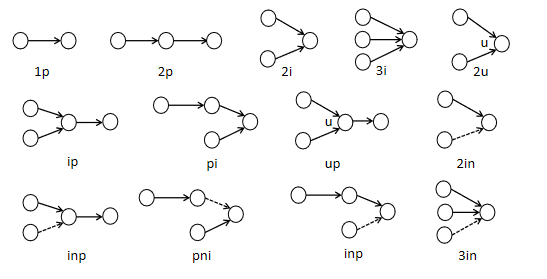}
    \vspace{-8pt}
    \caption{13 logical patterns.}
\end{wrapfigure}
The 13 predefined logical patterns used for hypothesis sampling are shown on the right.
Following~\citet{ctrlhgen}, five types of conditions are extracted from each query during data processing: \texttt{entitynumber}, the number of entities in the query; \texttt{relationnumber}, the number of relations in the query; \texttt{entity}, a specific entity in the query; \texttt{relation}, a specific relation in the query; and \texttt{pattern}, the corresponding e/p query pattern.

\paragraph{Multi-turn dialogues.}
For the multi-turn setting, we construct three-turn dialogues for each dataset. We first group queries by answer set and remove duplicate queries within each group. We then organize three queries from each group into a dialogue, with one query per turn.

The follow-up request at each turn follows one of three intention modes: \texttt{specific-condition}, \texttt{scope-narrowing}, and \texttt{scope-expansion}. The first turn always uses \texttt{specific-condition}, since there is no preceding query to compare against. From the second turn on, we compare the numbers of entities and relations in the current query with those in the next target query and sample a valid intention mode. We say that query $A$ is at least as complex as query $B$ if $A$ has at least as many entities and at least as many relations as $B$, or if the total number of entities and relations in $A$ is at least that of $B$. If the current query is at least as complex as the target query, \texttt{scope-expansion} is disabled; if the target query is at least as complex as the current query, \texttt{scope-narrowing} is disabled.
Under \texttt{scope-narrowing}, the follow-up asks to reduce the logical complexity, e.g., ``This is too complex. I want to make the logic simpler''; under \texttt{scope-expansion}, it asks to broaden the hypothesis, e.g., ``I want to know more. Let's explore more.''
We randomly select 500 dialogues from each dataset as test samples.
These dialogues are scripted from reference hypotheses rather than collected from real users. The labels \texttt{scope-narrowing} and \texttt{scope-expansion} describe changes in structural complexity and do not imply subset or superset relations between answer sets. Since a relative request may admit several plausible responses, the history comparison measures whether access to prior hypotheses helps the system reach the scripted target, rather than general dialogue understanding.

\section{Experiment Details}
\label{app:prompt}

\subsection{Lightweight Hypothesis Generation Model Training Details}

The lightweight hypothesis generator is a 6-layer Transformer trained with the AdamW optimizer.
The learning rate is set to \(1\times10^{-4}\), and the weight decay is set to \(1\times10^{-5}\).
For DBpedia50, we train the model for 200 epochs in the unconditional setting and 100 epochs in the conditional setting.
For BioKG, we train for 40 epochs in the unconditional setting and 15 epochs in the conditional setting.
For PharmKG8k, we train for 100 epochs in the unconditional setting and 160 epochs in the conditional setting.

\subsection{Inference Budget Accounting}
\label{app:budget}

All counts in Table~\ref{tab:budget} are averages per evaluated case. Each remote LLM request counts as one API call, including every request made within an RCA invocation. Generations count invocations of the hypothesis generator. KG executions count tool-level operations: a \texttt{graph\_validation} call counts once even if it executes multiple sub-queries, and a neighborhood search, including its internal candidate scoring, counts as one separate operation; primitive graph queries and individual candidate evaluations inside a tool are not counted. The table reports usage under each method's configured budget rather than under matched total computation. Likewise, best-of-$k$ sets $k$ to the maximum number of refinement rounds and does not match the total number of candidates evaluated by \modelname{}.

\subsection{Fragment Diagnosis Semantics}
\label{app:diagnosis}

Fragment diagnosis reports answer counts and overlap counts, whereas final candidate validation uses Jaccard similarity. The Refiner prompts prioritize low-overlap branches for inspection as a search heuristic. How these statistics should be read depends on the logical context: a conjunctive branch may return many extra entities that another branch filters out, a disjunctive branch need not cover all observations by itself, and an intermediate projection may return entities that play a different semantic role from the final target. Fragment statistics therefore suggest where to revise, while every proposed repair is evaluated as a complete hypothesis on the training graph.

\subsection{Inference Conditions and Candidate Roles}
\label{app:inference}

Within each dialogue turn, let $\mathbf{c}^{\mathrm{user}}$ denote the conditions extracted from the current request (resolved with conversational context), and let $\mathbf{c}_t$ denote the proposal conditions at refinement round $t$. The former defines the condition-adherence criterion for candidate selection, while the latter may include additional anchors suggested by the analysis. Both differ from the gold conditions used to compute Accuracy.

For conditional refinement, the Refiner prompt produces three candidate routes: regenerating with unchanged conditions, regenerating with updated conditions, and directly rewriting the hypothesis. Complete candidates are evaluated on the training graph and ranked first by adherence to $\mathbf{c}^{\mathrm{user}}$ and then by Jaccard. The best candidate is retained across rounds, and the final output is this best-so-far hypothesis, which is not necessarily the last round's proposal. The unchanged-condition route serves as a fallback when an updated condition is unhelpful.

For unconditional generation, there is no $\mathbf{c}^{\mathrm{user}}$. Instead, Prompt~3 induces structural, semantic, and hybrid hints from the initial query, and the resulting proposals are compared with the original unconditional output. These induced hints steer generation but are not enforced during selection, where candidates are ranked by training-graph Jaccard alone. An improved candidate may therefore use a different anchor from the one in the induced hint.

We use at most three refinement rounds and a training-graph Jaccard threshold of 0.95; RCA is triggered whenever the score of the current hypothesis falls below this threshold.

\subsection{Agent Details}

We report the prompts of the main LLM-based components of \modelname{}.
Fig.~\ref{fig:prompt-condition} shows the condition parsing prompt used by the Hypothesis Proposal Agent to convert user utterances into structured conditions.
Fig.~\ref{fig:prompt-rca} shows the Hypothesis Refiner Agent prompt for diagnosing and refining imperfect hypotheses.
Fig.~\ref{fig:prompt-uncondition} shows the prompt for inducing conditions from unconditional hypotheses in the unconditional setting.

\begin{figure*}[t]
\begin{AIbox}{Prompt 1: Hypothesis Proposal Agent}
{\bf Task:}\ You are the hypothesis proposal agent in knowledge graph abductive reasoning. Your task is to convert the user's conversation into the format suitable for the hypothesis generator, so as to use it to propose hypothesis later. Parse the following question into a JSON array of condition dicts.\\[2pt]
{\bf Valid condition types:}
\begin{itemize}
  \item \texttt{relation}: value = ONE relation NAME, e.g.\ \texttt{"GG"}
  \item \texttt{entity}: value = ONE entity NAME, e.g.\ \texttt{"chrnb3"}
  \item \texttt{relationnumber}: integer count of relations, e.g.\ \texttt{"3"}
  \item \texttt{entitynumber}: integer count of entities, e.g.\ \texttt{"2"}
  \item \texttt{pattern}: structural pattern using only \texttt{i/u/n/p/e} tokens, e.g.\ \texttt{"i p e p e"}
\end{itemize}
{\bf Rules:}
\begin{itemize}
  \item NEVER use \texttt{unconditional} — always infer at least one concrete condition.
  \item AT MOST ONE \texttt{relation} condition and AT MOST ONE \texttt{entity} condition.
  \item \texttt{relationnumber}/\texttt{entitynumber} take integer values, NOT names.
  \item \texttt{pattern} must use only \texttt{i/u/n/p/e} tokens.
  \item Never output empty string as value for \texttt{relation} or \texttt{entity}.
\end{itemize}
{\bf Examples:}\\
\texttt{"I want a hypothesis with relation GG"}\\
$\rightarrow$ \texttt{[\{"type":"relation","value":"GG"\}]}\\[2pt]
\texttt{"I want pattern i p e p e with 2 entities"}\\
$\rightarrow$ \texttt{[\{"type":"pattern","value":"i p e p e"\},}\\
\hspace*{1em}\texttt{\{"type":"entitynumber","value":"2"\}]}\\[4pt]
{\bf Question:} \textit{[user followup question]}\\[2pt]
{\bf Output:} Return ONLY the JSON array, nothing else.
\end{AIbox}
\caption{Hypothesis Proposal agent prompt used to convert natural language user requests into structured condition lists.}
\label{fig:prompt-condition}
\end{figure*}

\begin{figure*}[t]
\begin{AIbox}{Prompt 2: Hypothesis Refiner Agent}
{\bf Task:}\ You are an analysis agent for KG abductive reasoning. Given observed entities $O$, find a logical hypothesis $H$ such that executing $H$ on the KG returns exactly $O$. $H$ is one of 13 patterns (\texttt{1p/2p/2i/3i/ip/pi/2u/up/2in/3in/inp/pni/pin}). Analyze why the current hypothesis is imperfect, then propose 3 candidates to refine the hypothesis.\\[4pt]
{\bf Input:}
\begin{itemize}
  \item \textbf{Observations} (entity names): \textit{[answer\_nl]}
  \item \textbf{User's original condition} (must be respected): \textit{[original\_followup]}
  \item \textbf{Generation history}: \textit{[round, condition, jaccard per round]}
  \item \textbf{Current hypothesis}: NL form, raw action string, pattern, Jaccard score
  \item \textbf{ID$\leftrightarrow$Name lookup}: relation and entity id-to-name maps
\end{itemize}
\par\noindent\rule{\linewidth}{0.4pt}
{\bf Step 1 — Sub-logic Decomposition (Hypothesis Fragment Diagnosis)} (\texttt{graph\_validation})\\
Call \texttt{graph\_validation(query\_tokens, label\_answers, split='train')}. Each sub-query result contains \texttt{answer\_count}, \texttt{overlap\_count}, \texttt{relation\_to\_label}. Identify the weakest branch (lowest \texttt{overlap\_count}) to fix first. Tool budget: at most 3 calls.\\[4pt]
{\bf Step 2 — Neighborhood Search} (\texttt{incoming\_edge\_intersection})\\
Call \texttt{incoming\_edge\_intersection(answer\_entity\_ids, split='train', top\_k=10)}. Returns \texttt{flat\_candidates} (1-hop entity-relation pairs with Jaccard) and \texttt{two\_hop\_candidates} (2-hop paths). For \texttt{2i/3i/ip/pi} patterns, also call \texttt{intersection\_candidates(flat\_candidates\_json, mode)}.\\[4pt]
{\bf Step 3 — Produce 3 Candidates}\\
\textbf{Candidate 1} (keep): original condition unchanged, \texttt{hypothesis\_raw=null}.\\
\textbf{Candidate 2} (update): new condition extending original with analysis findings, \texttt{hypothesis\_raw=null}.\\
\textbf{Candidate 3} (generate): directly compose a flat action string \texttt{hypothesis\_raw} from tool results.\\
Flat format: space-separated tokens, no parentheses. Relations: negative integers (\texttt{-8}). Entities: positive integers (\texttt{1312}). E.g.\ \texttt{2i}: \texttt{"i -8 1312 -20 1303"}.\\[4pt]
{\bf Output:} \texttt{final\_answer(candidates)} — a list of 3 dicts with keys \texttt{analysis}, \texttt{new\_condition}, \texttt{hypothesis\_raw}.
\end{AIbox}
\caption{Hypothesis Refiner agent prompt used in the hypothesis refinement loop.}
\label{fig:prompt-rca}
\end{figure*}

\begin{figure*}[t]
\begin{AIbox}{Prompt 3: Unconditional Condition Generation Prompt}
{\bf Task:}\ An unconditional hypothesis was generated. Analyze it and generate structural and semantic conditions to improve hypothesis quality in subsequent conditional generation.\\[4pt]
{\bf Input:}
\begin{itemize}
  \item \textbf{Observations} (entity names): \textit{[answer\_nl]}
  \item \textbf{Unconditional hypothesis}: NL form, raw action string, Jaccard score
  \item \textbf{ID$\leftrightarrow$Name lookup}: relation id-to-name map
\end{itemize}
\par\noindent\rule{\linewidth}{0.4pt}
{\bf Step 1 — Sub-logic Decomposition((Hypothesis Fragment Diagnosis)} (\texttt{graph\_validation})\\
Call \texttt{graph\_validation} once. Use \texttt{overlap\_count} per branch to identify the strongest building block (semantic anchor) and the weakest branch (structural redesign target). Count sub-queries to infer \texttt{entitynumber} and \texttt{relationnumber}.\\[4pt]
{\bf Step 2 — Neighborhood Search} (\texttt{incoming\_edge\_intersection})\\
Call once with \texttt{top\_k=10}. Use top \texttt{flat\_candidates} to identify the best semantic (relation, entity) pair. Use \texttt{two\_hop\_candidates} to infer a good structural pattern (e.g.\ \texttt{2p} if top two-hop has high Jaccard).\\[4pt]
{\bf Step 3 — Return Condition Analysis}\\
Return a JSON object with three keys via \texttt{final\_answer()}:
\begin{itemize}
  \item \textbf{structural}: at least one of \texttt{entitynumber} (int 1--3), \texttt{relationnumber} (int 1--3), \texttt{pattern} (\texttt{i/u/n/p/e} tokens)
  \item \textbf{semantic}: at least one of \texttt{relation} (name string), \texttt{entity} (name string)
  \item \textbf{hybrid}: at least one structural key \textbf{and} at least one semantic key
\end{itemize}
Anchor conditions to the unconditional hypothesis already shown — reuse relations, entities, and structure already present rather than proposing unrelated constraints.
\end{AIbox}
\caption{Unconditional condition generation prompt: the agent analyzes an unconditional hypothesis and produces structural, semantic, and hybrid conditions for subsequent conditional generation.}
\label{fig:prompt-uncondition}
\end{figure*}

\subsection{Ablation Study}
\label{app:ablation}
Table~\ref{tab:singleturn_ablation} reports the detailed single-turn ablation results.
\begin{table}[t]
    \centering
    \scriptsize
    \caption{Single-turn ablation study.}
    \label{tab:singleturn_ablation}
    \setlength{\tabcolsep}{4pt}
    \renewcommand{\arraystretch}{1.1}
    \begin{tabular}{lccc}
    \toprule
    \textbf{Ablation} & \textbf{BioKG} & \textbf{PharmKG8k} & \textbf{DBpedia50} \\
    \midrule
    Hypoagent & 0.8846 & 0.8264 & 0.9445 \\
    w/o fragment & 0.8757  & 0.8128  & 0.9146  \\
    w/o neighborhood & 0.8754 & 0.7943 & 0.8973  \\
    w/o RCA& 0.7390 & 0.6439  & 0.8263 \\
    \bottomrule
    \end{tabular}
\end{table}

\section{Case Study}
\label{app:case}
\subsection{Single Turn}
Figure~\ref{fig:case-singleturn} illustrates how RCA and the Hypothesis Refiner Agent improve hypothesis generation in a single-turn PharmKG8k case.
Although the initial hypothesis satisfies the parsed structural constraints (two entities and two relations), its anchor entities only partially explain the observed drugs, resulting in low semantic similarity.
Condition satisfaction alone is thus insufficient: the model must also identify entities that are relevant to the observation set.

The Refiner revises the hypothesis by diagnosing weak anchors and exploring the KG neighborhood.
In this example, regenerating under the original condition does not improve the result.
In contrast, the updated condition (Candidate~2) introduces DB11914, which neighborhood search identifies as a more informative entity, in place of the weaker anchor of the initial hypothesis.
This targeted update yields a hypothesis that perfectly matches the observations.
Interestingly, directly combining the best neighborhood candidate with the stronger original anchor performs poorly, indicating that effective refinement requires not only retrieving promising entities but also composing them with suitable partners.
Overall, this case shows that RCA turns error diagnosis into actionable condition updates, enabling the generator to revise problematic fragments rather than merely regenerating from the original instruction.
\subsection{Multi-turn}
Fig.~\ref{fig:case-multiturn} presents a multi-turn case study on BioKG.
The observations are Reactome pathway or reaction entities, as indicated by their \texttt{R-HSA} identifiers.
This case shows how \modelname{} progressively adapts the generated hypothesis to evolving user intents across dialogue turns.

In Turn~1, the user requests a hypothesis with three entities and the relation \textit{MEMBER\_OF\_COMPLEX}.
The Proposal Agent parses these constraints, and the system generates a relatively complex hypothesis involving negation and a protein--protein interaction path. Its description names only two anchors rather than the three requested, so the entity-number condition may not be fully satisfied.
Although this hypothesis already achieves high semantic similarity, its complexity conflicts with the user's subsequent preference for simplicity.
In Turn~2, the user states that the hypothesis is too complex and asks for simpler logic.
The system interprets this relative request as reducing the number of entities while keeping the core relation \textit{MEMBER\_OF\_COMPLEX}, producing a simpler hypothesis with higher semantic similarity.
This is one reasonable interpretation of a simplification request among several possible ones.

In Turn~3, the system interprets ``know more and explore more'' as a request for broader coverage rather than further simplification.
It accordingly changes the pattern to a two-way union and generates a disjunctive hypothesis over two \textit{MEMBER\_OF\_COMPLEX} anchors.
This final hypothesis perfectly matches the observation set, showing that \modelname{} can adjust both the structural pattern and the anchor entities during multi-turn interaction.
Overall, this case shows how the query structure evolves across successive requests, and also illustrates that answer-set quality and condition adherence are distinct criteria.

\subsection{Unconditional}
Fig.~\ref{fig:case-uncondition} shows an unconditional generation case on DBpedia50.
The observations are football players, most of whom appear in professional football transfer records.
Without any user-provided condition, the initial hypothesis links the observations to the entity
\textit{List\_of\_Iranian\_football\_transfers\_summer\_2012} through the relation \textit{currentMember}, but covers only a small portion of the observation set.

The agent then induces structural, semantic, and hybrid conditions from the initial hypothesis.
The structural condition preserves only the logical form, i.e., the numbers of entities and relations and the path pattern.
Since it leaves the key semantic anchor unchanged, the generated hypothesis is identical to the original one.
Similarly, the hybrid condition still constrains generation to the original structure and relation, and thus fails to escape the initial local solution.
In contrast, the semantic condition focuses on the relation \textit{currentMember} and the transfer-list entity, allowing the model to search within a semantically related neighborhood.
As a result, the model finds a closely related transfer list,
\textit{List\_of\_Iranian\_football\_transfers\_winter\_2011\text{--}12},
which better explains the observed players.

Note that the improved candidate replaces the summer-2012 anchor named in the induced hint with the winter-2011--12 list, and thus does not strictly satisfy that hint. This is acceptable in the unconditional setting, where induced hints steer generation but are not enforced during selection. Moreover, a single case does not imply that semantic hints are generally more effective than structural ones.

\begin{figure*}[t]
\begin{AIbox}{Case Study: Single-turn Hypothesis Generation with RCA}
{\bf Observations:} DB11988, DB15253, DB12489, DB14724, DB13952, DB13953, DB13954, DB12530, DB13956, DB12089, DB15595, DB14762, DB14919\\
{\bf User Condition:} I want a hypothesis that contains 2 entities and has 2 relations.\\
{\bf Parsed Conditions:} \texttt{entitynumber=2}, \texttt{relationnumber=2}\\[4pt]
\par\noindent\rule{\linewidth}{0.4pt}
{\bf Initial Generation}\\
{\bf Hypothesis:} $H = V_?: \textit{DDI}(\textit{DB00163}, V_?) \;\wedge\; \textit{DDI}(\textit{DB11972}, V_?)$\\
{\bf NL:} Drugs that have a drug-drug interaction with both DB00163 and DB11972.\\
{\bf Jaccard:} 0.231\\[4pt]
\par\noindent\rule{\linewidth}{0.4pt}
{\bf Hypothesis Refiner Agent — update the condition} \\
{\bf Condition:} I want a hypothesis that contains 2 entities and has 2 relations and includes entity DB11914.\\
{\bf Analysis:} Neighborhood search identifies DB11914 as the top flat candidate (Jaccard=0.097 via DDI). Replacing the weaker anchor DB00163 with DB11914  should improve coverage.\\
{\bf Hypothesis:} $H = V_?: \textit{DDI}(\textit{DB11914}, V_?) \;\wedge\; \textit{DDI}(\textit{DB01590}, V_?)$\\
{\bf NL:} Drugs that have a drug-drug interaction with both DB11914 and DB01590.\\
{\bf Jaccard:} \textbf{1.000}\\[4pt]
\par\noindent\rule{\linewidth}{0.4pt}
{\bf Hypothesis Refiner Agent — directly refine the hypothesis}\\
{\bf Analysis:} Compose a 2i directly using the best flat candidate DB11914 and the stronger original anchor DB11972, both via DDI.\\
{\bf Hypothesis:} $H = V_?: \textit{DDI}(\textit{DB11914}, V_?) \;\wedge\; \textit{DDI}(\textit{DB11972}, V_?)$\\
{\bf NL:} Drugs that have a drug-drug interaction with both DB11914 and DB11972.\\
{\bf Jaccard:} 0.083
\end{AIbox}
\caption{Single-turn case study with Root Cause Analysis.}
\label{fig:case-singleturn}
\end{figure*}

\begin{figure*}[t]
\begin{AIbox}{Case Study: Multi-turn Hypothesis Generation}
{\bf Observations:} R-HSA-983412, R-HSA-983413, R-HSA-983414, R-HSA-983416, R-HSA-983418, R-HSA-983419, R-HSA-983420, R-HSA-198896, R-HSA-198897, R-HSA-198904, R-HSA-198906, R-HSA-8951496, R-HSA-1236916, R-HSA-1236922, R-HSA-1236928, R-HSA-167755, R-HSA-1236930, R-HSA-1236931, R-HSA-1236934, R-HSA-983114, R-HSA-983116, R-HSA-983119, R-HSA-983120, R-HSA-983121, R-HSA-199592, R-HSA-983127, R-HSA-983132, R-HSA-983136, R-HSA-8863933\\[4pt]
\par\noindent\rule{\linewidth}{0.4pt}
{\bf Turn 1}\\
{\bf User Condition:} I want a hypothesis that contains 3 entities and includes the relation ``MEMBER\_OF\_COMPLEX''.\\
{\bf Parsed Conditions:} \texttt{relation=MEMBER\_OF\_COMPLEX}, \texttt{entitynumber=3}\\
{\bf Query structure:} A negated membership branch intersected with a two-relation path. The description below names two anchors, not the three requested.\\
{\bf NL:} Entities that are \emph{not} members of the complex containing O95971, but are reachable via PPI from P30511 through MEMBER\_OF\_COMPLEX.\\
{\bf Jaccard:} 0.800
\par\noindent\rule{\linewidth}{0.4pt}
{\bf Turn 2}\\
{\bf User Condition:} This is too complex. I want to make the logic simpler.\\
{\bf Parsed Conditions:} \texttt{relation=MEMBER\_OF\_COMPLEX}, \texttt{entitynumber=1}\\
{\bf Query structure:} A two-relation projection path with one named anchor.\\
{\bf NL:} Entities reachable via PPI from P01889 through MEMBER\_OF\_COMPLEX.\\
{\bf Jaccard:} 0.829
\par\noindent\rule{\linewidth}{0.4pt}
{\bf Turn 3}\\
{\bf User Condition:} I want to know more and explore more.\\
{\bf Parsed Conditions:} \texttt{pattern=2u}\\
{\bf Best Hypothesis:}
$H = V_?: \textit{MEMBER\_OF\_COMPLEX}(\textit{P01893}, V_?) \;\vee\; \textit{MEMBER\_OF\_COMPLEX}(\textit{P10321}, V_?)$\\
{\bf NL:} Entities that are members of the complex containing P01893 \emph{or} P10321.\\
{\bf Jaccard:} \textbf{1.000}
\end{AIbox}
\caption{Multi-turn case study. Path compositions are described in natural language. A high semantic score does not imply that every requested condition is satisfied.}
\label{fig:case-multiturn}
\end{figure*}

\begin{figure*}[t]
\begin{AIbox}{Case Study: Unconditional Generation with Condition Induction}
For compact notation, let $e_s$ denote \textit{List\_of\_Iranian\_football\_transfers\_summer\_2012} and $e_w$ denote \textit{List\_of\_Iranian\_football\_transfers\_winter\_2011\text{--}12}.\\[4pt]
{\bf Observations:} Mohammad\_Mansouri, Giorgi\_Krasovski, Yaroslav\_Krushelnitskiy, Nenad\_Brnović, Mohsen\_Hamidi\\[4pt]
\par\noindent\rule{\linewidth}{0.4pt}
{\bf Unconditional Hypothesis}\\
{\bf Hypothesis:} $H(X) = \textit{currentMember}(e_s, X)$\\
{\bf NL:} Entities that are current members of the Iranian football transfers list (summer 2012).\\
{\bf Jaccard:} 0.208\\[4pt]
\par\noindent\rule{\linewidth}{0.4pt}
{\bf Candidate 1 — Unconditional}\\
{\bf Conditions:} (none)\\
{\bf Hypothesis:} $H(X) = \textit{currentMember}(e_s, X)$\\
{\bf NL:} Entities that are current members of the Iranian football transfers list (summer 2012).\\
{\bf Jaccard:} 0.208\\[4pt]
\par\noindent\rule{\linewidth}{0.4pt}
{\bf Candidate 2 — Structural}\\
{\bf Conditions:} \texttt{entitynumber=1}, \texttt{relationnumber=1}, \texttt{pattern=p e}\\
{\bf Hypothesis:} $H(X) = \textit{currentMember}(e_s, X)$\\
{\bf NL:} Entities that are current members of the Iranian football transfers list (summer 2012).\\
{\bf Jaccard:} 0.208\\[4pt]
\par\noindent\rule{\linewidth}{0.4pt}
{\bf Candidate 3 — Semantic}\\
{\bf Conditions:} \texttt{relation=currentMember}, \\ \quad \texttt{entity=List\_of\_Iranian\_football\_transfers\_summer\_2012}\\
{\bf Hypothesis:} $H(X) = \textit{currentMember}(e_w, X)$\\
{\bf NL:} Entities that are current members of the Iranian football transfers list (winter 2011--12).\\
{\bf Jaccard:} \textbf{0.833}\\[4pt]
\par\noindent\rule{\linewidth}{0.4pt}
{\bf Candidate 4 — Hybrid}\\
{\bf Conditions:} \texttt{entitynumber=1}, \texttt{relation=currentMember}, \texttt{pattern=p e}\\
{\bf Hypothesis:} $H(X) = \textit{currentMember}(e_s, X)$\\
{\bf NL:} Entities that are current members of the Iranian football transfers list (summer 2012).\\
{\bf Jaccard:} 0.208
\end{AIbox}
\caption{Unconditional generation case study.}
\label{fig:case-uncondition}
\end{figure*}

\section{Standard Deviation Tables}
\label{app:variance}

Table~\ref{tab:singleturn_std} reports the standard deviations for the single-turn results in Table~\ref{tab:singleturn}, and Table~\ref{tab:multiturn_std} reports those for the multi-turn results in Table~\ref{tab:multiturn}.
These are population standard deviations across evaluated cases (divisor $N$, i.e., \texttt{np.std} with \texttt{ddof=0}) rather than variability across repeated runs. All metrics lie in $[0,1]$. For binary condition adherence, the standard deviation is $\sqrt{p(1-p)}$, where $p$ is the mean accuracy; since these entries are computed from the rounded means in the main tables, they are approximate.

\begin{table*}[t]
    \centering
    \scriptsize
    \caption{Standard-deviation values for single-turn controllable hypothesis generation.}
    \label{tab:singleturn_std}
    \setlength{\tabcolsep}{2.6pt}
    \renewcommand{\arraystretch}{1.15}
    \begin{tabular}{lcccccccccccc}
    \toprule
    \multirow{2}{*}{\textbf{Model}} & \multicolumn{4}{c}{\textbf{BioKG}} & \multicolumn{4}{c}{\textbf{PharmKG8k}} & \multicolumn{4}{c}{\textbf{DBpedia50}} \\
    \cmidrule(lr){2-5} \cmidrule(lr){6-9} \cmidrule(lr){10-13}
    & \textbf{Jaccard} & \textbf{Dice} & \textbf{Overlap} & \textbf{Accuracy}
    & \textbf{Jaccard} & \textbf{Dice} & \textbf{Overlap} & \textbf{Accuracy}
    & \textbf{Jaccard} & \textbf{Dice} & \textbf{Overlap} & \textbf{Accuracy} \\
    \midrule
    Symbolic Search
    & 0.40 & 0.41 & 0.46 & --
    & 0.35 & 0.34 & 0.35 & --
    & 0.46 & 0.47 & 0.50 & -- \\
    ReAct
    & 0.38 & 0.39 & 0.46 & 0.447
    & 0.36 & 0.37 & 0.44 & 0.465
    & 0.47 & 0.48 & 0.49 & 0.496 \\
    \midrule
    CtrlHGen
    & 0.37 & 0.35 & 0.32 & 0.288
    & 0.36 & 0.33 & 0.31 & 0.263
    & 0.31 & 0.28 & 0.26 & 0.444 \\
    CtrlHGen (best-of-$k$)
    & 0.34 & 0.31 & 0.27 & 0.295
    & 0.32 & 0.28 & 0.23 & 0.262
    & 0.30 & 0.27 & 0.25 & 0.454 \\
    ReAct + CtrlHGen
    & 0.36 & 0.34 & 0.28 & 0.398
    & 0.40 & 0.38 & 0.36 & 0.431
    & 0.29 & 0.26 & 0.23 & 0.468 \\
    \midrule
    \modelname{}$_{\textsc{DS}}$
    & 0.22 & 0.18 & 0.13 & 0.236
    & 0.25 & 0.20 & 0.13 & 0.247
    & 0.16 & 0.13 & 0.10 & 0.435 \\
    \modelname{}$_{\textsc{Qwen}}$
    & 0.24 & 0.20 & 0.14 & 0.237
    & 0.25 & 0.20 & 0.15 & 0.230
    & 0.20 & 0.15 & 0.11 & 0.443 \\
    \modelname{}$_{\textsc{GPT}}$
    & 0.22 & 0.18 & 0.13 & 0.228
    & 0.25 & 0.20 & 0.13 & 0.232
    & 0.16 & 0.13 & 0.10 & 0.441 \\
    \bottomrule
    \end{tabular}
\end{table*}

\begin{table*}[t]
    \centering
    \scriptsize
    \caption{Standard-deviation values for multi-turn hypothesis generation.}
    \label{tab:multiturn_std}
    \setlength{\tabcolsep}{2.2pt}
    \renewcommand{\arraystretch}{1.15}
    \begin{tabular}{lcccccccccccc}
    \toprule
    \multirow{2}{*}{\textbf{Method}}
    & \multicolumn{4}{c}{\textbf{BioKG}} & \multicolumn{4}{c}{\textbf{PharmKG8k}} & \multicolumn{4}{c}{\textbf{DBpedia50}} \\
    \cmidrule(lr){2-5} \cmidrule(lr){6-9} \cmidrule(lr){10-13}
    & \textbf{Jaccard} & \textbf{Dice} & \textbf{Overlap} & \textbf{Accuracy}
    & \textbf{Jaccard} & \textbf{Dice} & \textbf{Overlap} & \textbf{Accuracy}
    & \textbf{Jaccard} & \textbf{Dice} & \textbf{Overlap} & \textbf{Accuracy} \\
    \midrule
    ReAct
    & 0.45 & 0.45 & 0.49 & 0.222
    & 0.42 & 0.43 & 0.45 & 0.241
    & 0.48 & 0.48 & 0.49 & 0.297 \\
    \midrule
    CtrlHGen (best-of-$k$)
    & 0.39 & 0.39 & 0.37 & 0.428
    & 0.29 & 0.26 & 0.23 & 0.423
    & 0.34 & 0.32 & 0.32 & 0.472 \\
    ReAct + CtrlHGen
    & 0.47 & 0.47 & 0.49 & 0.387
    & 0.44 & 0.44 & 0.43 & 0.419
    & 0.43 & 0.48 & 0.48 & 0.432 \\
    \modelname{} (w/o history)
    & 0.26 & 0.24 & 0.20 & 0.360
    & 0.19 & 0.18 & 0.12 & 0.395
    & 0.23 & 0.20 & 0.18 & 0.456 \\
    \modelname{} (w/ history)
    & 0.24 & 0.22 & 0.19 & 0.369
    & 0.18 & 0.16 & 0.12 & 0.396
    & 0.21 & 0.18 & 0.15 & 0.438 \\
    \bottomrule
    \end{tabular}
\end{table*}

\section{Post-hoc Candidate Removal Analysis}
\label{app:candidate-ablation}

To examine the contribution of the Hypothesis Refiner Agent's candidate strategies, we remove one candidate category from the pools recorded during full-system runs and reselect the output. We distinguish \textbf{condition updates} (Candidate~2 in Fig.~\ref{fig:prompt-rca}) from \textbf{direct rewriting} (Candidate~3); regeneration under unchanged conditions (Candidate~1) remains available in both settings. Reselection follows the original rule, prioritizing the original user conditions and then training-graph Jaccard, and only the selected hypothesis is evaluated on the test graph.
Since this analysis is post hoc rather than a rerun of the loop with a strategy disabled, the retained candidates may still depend on the history produced by the full system.

Table~\ref{tab:candidate_ablation_jaccard} reports the average test-graph Jaccard for the full candidate pools and the two removal settings.
Removing condition-update candidates causes larger average decreases on all datasets (0.148, 0.132, and 0.128) than removing directly rewritten candidates (0.048, 0.032, and 0.091).
Condition updates therefore contribute more to average quality, while direct rewriting provides complementary candidates.

Table~\ref{tab:candidate_ablation_samples} reports the percentage of evaluated samples whose test-graph Jaccard decreases relative to the full system after reselection. Removing condition-update candidates affects 33.6--53.0\% of samples, compared with 14.0--24.0\% for direct rewriting. Note that these are degradation rates rather than win rates.

\begin{table}[t]
    \centering
    \scriptsize
    \caption{Average test-graph Jaccard after post-hoc candidate removal and training-graph-based reselection.}
    \label{tab:candidate_ablation_jaccard}
    \setlength{\tabcolsep}{6pt}
    \renewcommand{\arraystretch}{1.1}
    \begin{tabular}{lccc}
    \toprule
    \textbf{Setting} & \textbf{BioKG} & \textbf{PharmKG8k} & \textbf{DBpedia50} \\
    \midrule
    Full system & 0.885 & 0.826 & 0.945 \\
    w/o condition updates & 0.737 & 0.694 & 0.817 \\
    w/o direct rewriting & 0.837 & 0.794 & 0.854\\
    \bottomrule
    \end{tabular}
\end{table}

\begin{table}[t]
    \centering
    \scriptsize
    \caption{Percentage of evaluated samples with lower test-graph Jaccard after post-hoc candidate removal, relative to the full system.}
    \label{tab:candidate_ablation_samples}
    \setlength{\tabcolsep}{6pt}
    \renewcommand{\arraystretch}{1.1}
    \begin{tabular}{lccc}
    \toprule
    \textbf{Removed category} & \textbf{BioKG} & \textbf{PharmKG8k} & \textbf{DBpedia50} \\
    \midrule
    Condition updates & 44.2\% & 53.0\%& 33.6\% \\
    Direct rewriting & 14.0\% & 15.6\% & 24.0\%  \\
    \bottomrule
    \end{tabular}
\end{table}

\section{AI use statement}

We used generative AI tools to polish the language of the manuscript, assist with writing, and help implement experimental and analysis code. We have reviewed all AI-assisted text and code, and take full responsibility for the final content of this paper, including its claims, wording, and experimental results.

\end{document}